\documentclass[journal abbreviation, manuscript]{copernicus}

\usepackage{tabularx}



\usepackage{siunitx}

\begin{document}
\nolinenumbers   

\title{Detecting micro fractures: A comprehensive comparison of conventional and machine-learning based segmentation methods}


\Author[1]{Dongwon}{Lee}
\Author[1]{Nikolaos}{Karadimitriou}
\Author[1]{Matthias}{Ruf}
\Author[1,2]{Holger}{Steeb}

\affil[1]{Institute of Applied Mechanics (CE), Pfaffenwaldring 7, University of Stuttgart,70569, Germany}
\affil[2]{Stuttgart Center for Simulation Science, Pfaffenwaldring 5a, University of Stuttgart,70569, Germany}




\correspondence{Dongwon Lee (dongwon.lee@mechbau.uni-stuttgart.de)}

\runningtitle{Detecting micro fractures: A comprehensive comparison of
conventional and machine learning based segmentation methods}

\runningauthor{Dongwon Lee, Nikolaos Karadimitriou, Matthias Ruf, Holger Steeb}

\received{}
\pubdiscuss{} 
\revised{}
\accepted{}
\published{}


\firstpage{1}

\maketitle

\begin{abstract}
Studying porous rocks with X-Ray Computed Tomography (XRCT) has been established as a standard procedure for the non-destructive characterization of flow and transport in porous media. Despite the recent advances in the field of XRCT, various challenges still remain due to the inherent noise and imaging artefacts in the produced data. These issues become even more profound when the objective is the identification of fractures, and/or fracture networks. One challenge is the limited contrast between the regions of interest and the neighboring areas, which can mostly be attributed to the minute aperture of the fractures. In order to overcome this challenge, it has been a common approach to apply various digital image processing steps, such as filtering, to enhance the signal-to-noise ratio. Additionally, segmentation methods based on threshold/morphology schemes have been employed to obtain enhanced information from the features of interest. However, this workflow needs a skillful operator to fine-tune its input parameters, and the required computation time significantly increases due to the complexity of the available methods, and the large volume of an XRCT data-set. In this study, based on a data-set produced by the successful visualization of a fracture network in Carrara marble with $\mu$XRCT, we present the results from five segmentation methods, three conventional and two machine learning-based ones. The objective is to provide the interested reader with a comprehensive comparison between existing approaches, while presenting the operating principles, advantages and limitations, to serve as a guide towards an individualized segmentation workflow. The segmentation results from all five methods are compared to each other in terms of quality and time efficiency. Due to memory limitations, and in order to accomplish a fair comparison, all the methods are employed in a 2D scheme. The output of the 2D U-net model, which is one of the adopted machine learning-based segmentation methods, shows the best performance regarding the quality of segmentation and the required processing time. 
\end{abstract}

\introduction\label{sec:introduction}  
The adequate characterization and knowledge of the geometrical properties of fractures has a strong constructive impact on various disciplines in engineering and geosciences. In the field of reservoir management, due to the fact that fractures strongly influence the permeability and porosity of a reservoir, their properties, like network connectivity and mean aperture, play a decisive role \cite{Berkowitz1995,Jing2017,Karimpouli2017, SUZUKI} in the production and conservation processes of resources, such as methane \cite{Jiang2016} and shale oil \cite{Dong2019, VANSANTVOORT201828}. Additionally, the geometry of the fracture network is an equally important property that needs to be addressed in our efforts to properly characterize the stresses developing between tectonic plates \cite{Lei2018} which, eventually, have a strong effect on the permeability of a reservoir \cite{Crawford2017,Huang2019}. As one of the most straight-forward and effective ways of investigating such fracture properties, 3D visualization and consequent efficient extraction of various fractures, or fracture network, spatial properties, like length, aperture, tortuosity etc., has been increasingly employed.

X-Ray Computed Tomography (XRCT) has been adopted as a non-destructive method to monitor and quantify such spatial properties, and it has become a standard procedure in laboratory studies \cite{Weerakone2010,Taylor2015, DeKock2015,Saenger2016,Halisch2016,Chung2019}. More specifically, lab-based micro-X-Ray Computed Tomography ({$\mu$XRCT}) offers the ability to visualize features on the pore-scale at micrometer resolution.
However, as with every other visualization method, there are some drawbacks which are inherent to the method itself. This visualization method is based on the interaction of an, otherwise, opaque sample with a specific type of radiation. Occasionally, there is a low contrast between distinct features, either belonging to different materials, or having spatial characteristics which are close to the resolution of the method itself. Such an example is the distinction between the solid matrix and an existing fracture, which commonly has a very small aperture \cite{Lai2017,Kumari2018,Zhang2018,Furat2019, ALZUBAIDI2022109471}. Consequently, a lot of effort has to be made to ``clean'' the acquired data from potential noise, independently of its source \cite{Frangakis2001,Behrenbruch2004,Christe2007}. As a means to tackle this issue, there have been many studies focusing on the enhancement and optimization of digital image processing with the use of filtering methods \cite{Behrenbruch2004,Coady2019} which could reduce the inherent noise of the image-based data while keeping the features of interest intact. In this way ``clean'' images, meaning with a low signal-to-noise ratio, can be obtained with an enhanced contrast between the features of interest and their surroundings. 

In the work of \cite{Poulose2013}, the authors classified the available filtering methods. Their classification can be narrowed down into two major categories, namely the linear and the non-linear ones \cite{Ahamed2019}. The latter type of filtering methods makes use of the intensities of the neighboring pixels in comparison to the intensities of the ones of interest. There are two well-known filtering methods of this type; the first one is commonly referred to as ``anisotropic diffusion'' \cite{Perona1990},  where a blurring scheme is applied in correlation to a threshold gray-scale value. The other is commonly referred to as the ``non-local means filter'' \cite{Buades2011}. This method uses mean intensities to compute the variance (similarity) at the local area surrounding a target pixel. As an improvement to this filter, the ``adaptive manifolds non-local means filter'' \cite{Gastal2012} has been recently developed to reduce the data noise with the help of an Euclidean-geodesic relation between pixel intensities and the calculated local mean intensities of the created manifolds, accounting for the grayscale value as a separate dimension for each pixel. On the other hand, the linear type of filtering method makes use only of fixed (pixel-oriented) criteria. This, in practice, means that it simply blurs the image without distinguishing between generic noise and the boundaries of an object of interest. Although there are no prerequisites in the selection of the appropriate noise filtering method, the non-linear type of filtering is more frequently employed with XRCT due to its edge conserving characteristic \cite{Sheppard2004,Taylor2015,Ramandi2017}.  

Conventionally, image segmentation can take place simply by applying a certain threshold to classify the image corresponding to pixel intensity. The most well-known approach based on this scheme is the ``Otsu  method'' \cite{NobuyukiOtsu1979}. In this segmentation scheme, histogram information from intensities is used to return a threshold value which classifies the pixels into two classes, one of interest and the rest. As an advanced intensity-based approach for multi-class applications, the K-mean method is often used \cite{Salman2006,Dhanachandra2015,Chauhan2016,macqueen1967}. \cite{Dhanachandra2015} showed their effectively segmented results of malaria-infected blood cells with the help of this method. However, this intensity-based segmentation approach could not be a general solution for XRCT data due to lab-based XRCT inherent ``beam-hardening'' effects, which creates imaging artefacts. These artifacts are caused by polychromatic X-rays (more than one wavelengths) which induce different attenuation rates along the thickness of a sample \cite{Ketcham2014}. As another common approach of segmentation, the methods based on ``edge-detection'' are often adopted \cite{Su2011,BHARODIYA2019e02743}. In this scheme one could utilize the contrast of intensities within the images which would eventually outline the features of interest \cite{Al-amri2010}. 
Based on the assumption that the biggest contrast would be detected where two differently gray-scaled features meet, there are many different types of operators in order to compute this contrast, namely the Canny \cite{Canny}, Sobel \cite{Sobel} and Laplacian \cite{Marr1980} ones. However, there is an inherent limitation to apply the method to XRCT applications, since the image has to have a high signal-to-noise ratio, which is hardly ever the case when fractures are involved. 

\cite{Voorn2013}, adopted a Hessian matrix-based scheme and employed ``Multi-scale Hessian Fracture'' (MHF) filtering to enhance the narrow planar shape of fractures in 3D XRCT data from rock samples, after applying a Gaussian filter to improve fracture visibility. This resulted in the enhancement of fractures of various apertures. The authors easily binarized their images with the help of emphasized responses with structural information obtained from the Hessian matrix. This scheme was also employed in the work of \cite{Deng2016}, where they introduced the ``Technique of Iterative Local Thresholding'' (TILT) in order to segment a single fracture in their rock image data-sets. After acquiring an initial image just to get a rough shape of the fracture by using a threshold or MHF, they applied the \textit{Local threshold} method \cite{localthreshold}, which sets a threshold individually for every pixel (cf. ~\ref{sec:local_threshold}), repeatedly to narrow down the rough shape into a finer one, until its boundary reached the outline of the fracture. Despite the successful segmentation of their data-set, some challenges still remained for the segmentation of fracture networks, as expressed in the work of \cite{Drechsler2010}.  The authors compared the famous Hessian-based algorithms from \cite{Sato1997}, \cite{Frangi1998}, and \cite{Erdt2008} with each other. In their study they showed that these schemes are not too effective to classify the intersections of the features of interest. Additionally, in the case of MHF, due to the used Gaussian filter which blurs the edges of a fracture, the surface roughness of the fractures would most probably not be identified. Also, the segmentation results varied significantly corresponding to the chosen range of standard deviations \cite{Deng2016}.

As a more advanced segmentation approach, the watershed segmentation method was introduced by \cite{Beucher1993}. It allows to obtain individually assigned groups of pixels (features) as a segmentation result. This can be accomplished by following the following steps:
\begin{enumerate}
    \item Find the local intensity minimums for each one of the assigned groups of pixels.
    \item At each local minimum, the assignment of the nearby pixels to individual label groups is performed according to the magnitudes of the corresponding intensity gradients. A distance map, where the distance is computed from all non-zero pixels of a binarized image, could be used instead of an image which contains intensity values \cite{Maurer2003}.
    \item The above mentioned procedures ought to be performed repetitively until there are no more pixels to be assigned.
\end{enumerate}
\cite{Taylor2015}, based on the watershed method, were able to successfully visualize a segmented void space of XRCT scanned granular media (sand) which also had a narrow and elongated shape, similar to a fracture. \cite{Ramandi2017} demonstrated the segmentation result of coal cleats by applying a more advanced scheme based on the watershed method, the Converging Active Contour (CAC) method, initially developed by \cite{Sheppard2004}. Their method combined two different schemes; one was the watershed method itself, and the other was the active contouring method \cite{Caselles1993}, which is commonly used for finding the continuous and compact outline of features. However, there was still a fundamental limitation to cope with, to directly segment a fracture with the introduced method since the method requires a well-selected initial group of pixels, while it is hard to define those only with a threshold due to the low contrast of the feature \cite{Weerakone2010}. The same authors \cite{Ramandi2017} applied a saturation technique with an X-ray attenuating fluid. By performing this they were able to enhance the visibility of the pores and fractures within the XRCT data.

Instead of dealing with the image with numerically designed schemes, in recent years and in various fields, like remote sensing \cite{Li2019,CRACKNELL}, medical imaging \cite{Singh2020} and computer vision \cite{Minaee2020}, image segmentation using machine learning techniques has been being robustly studied as a new approach due to its well-known benefits, like flexibility to all types of applications and less demanding user intervention. Based on this advantages, \cite{Kodym2018} introduced a semi-automatic image segmentation workflow with XRCT data by using the Random Forest algorithm \cite{Amit1997} which statistically evaluated the output from decision trees (cf.~\ref{sec:Random_forest}). \cite{Karimpouli2019} demonstrated cleats segmentation results from XRCT with a coal data sample by applying the Convolutional Neural Networks (CNNs) approach, which is another machine learning scheme. The CNN is one of the deep artificial neural network models which consists of multi-layers with convolutional and max-pooling layers \cite{Albawi2018, PALAFOX}. The former type of layers is employed to extract features out of an input image and the latter one is employed to down-sample the input by collecting the maximum coefficients which are obtained from the convolutional layers. Based on the CNNs architecture, \cite{Long2017} demonstrated the Fully Convolutional Networks (FCN) model. The model consists of an encoding part, which is of the same structure as the CNNs, and a decoding part, which up-samples the layers using a de-convolutional operation (cf.~\ref{sec:U-net}). The authors also introduced a skip connection scheme, which is concatenating the previously maximum collected layers into the decoding part in order to enhance the prediction of the model. 

Despite these promising results from the use of machine learning schemes, there is still room for improvement. In a machine learning-based scheme, apart from improving the architecture of the training model, which significantly affects the efficiency and output of the model \cite{Khryashchev2018}, one major issue is the provision of a ground-truth data-set. In most applications \cite{Shorten2019,Roberts2019,Zhou2019,Alqahtani2020} this was performed based on subjective manual annotation which would consume noticeable time and effort of an expert. This manual annotation is hardly the best approach, especially in our application which contains sophisticated shapes of fracture networks and low contrast features due to the low density of air within the small volume of a fracture \cite{Furat2019}. 

With the use of a data-set taken from the imaging of a real fractured porous medium, and for comparative purposes, three different segmentation techniques were adopted in this work, based on conventional means, such as the Sato, Local threshold, and active contouring method, including filtering and post-processing. Two different machine learning models were employed for this comparison as well. The well-known U-net \cite{Ronneberger2015}, as well as the Random Forest model, were evaluated, with the Random Forest model being supported in Trainable Weka \cite{ArgandaCarreras2017}. Due to memory limitation for the U-net model we had to dice the 2D images to smaller 2D tiles and compute sequentially. The computed prediction tiles were merged together at a later stage to reconstruct the 2D image. In order to train the adopted model, we provided a training data-set which was obtained by means of a conventional segmentation method. Note that only a small portion of the results (60 slices out of 2140 images) was used as a training data-set. For a fair comparison between methods, all segmentation schemes were applied on 2D data. 
Via comparison, we were able to identify that the U-net segmentation model outperformed the others in terms of quality of the output and time efficiency. Some defects which appeared in the provided truth data by conventional segmentation means were smeared out in the predictions of the model.

\section{Materials and methods}

\subsection{Carrara marble sample and data acquisition}

The segmentation approach presented here is based on a $\mu$XRCT data-set of a thermally treated cylindrical Bianco Carrara marble core sample with a diameter of \SI{5}{\milli\meter} and a length of \SI{10}{\milli\meter}. Bianco Carrara marble is a crystalline rock, consisting of about \SI{98}{\percent} calcite (CaCO$_3$) \cite{Pieri2001} and is a frequently used material in experimental rock mechanics \cite{DellePiane2015}. In combination with mechanical or thermal treatment, the virgin state can be modified to achieve different characteristics of micro-fractures within the sample \cite{Lissa2020,Lissa2021,Pimienta2019b,Sarout2017,DellePiane2015,Peacock1994}.

The considered sample was extracted from a bigger cylindrical core sample which was subjected to a thermal treatment beforehand. This treatment included:\hfill \break 1. Heating-up from room temperature (\SI{20}{\degree}C) to \SI{600} {\degree}C with a heating rate of \SI{3}{\kelvin\per\minute}.\hfill \break  2. Holding this temperature for \SI{2}{\hour} to ensure a uniform temperature distribution in the entire sample.\hfill \break  3. Quenching the sample in a water basin at room temperature (\SI{20}{\celsius}). \hfill \break The porosity of the raw material, before any treatment, was \SI{0.57}{\percent} measured with mercury porosimetry. The porosity of the thermally treated sample was obtained from measurements of volume changes before/after quenching, and was found to be around 3 \%. 

\begin{figure}[h]
    \centering
    \includegraphics[width=0.5\textwidth]{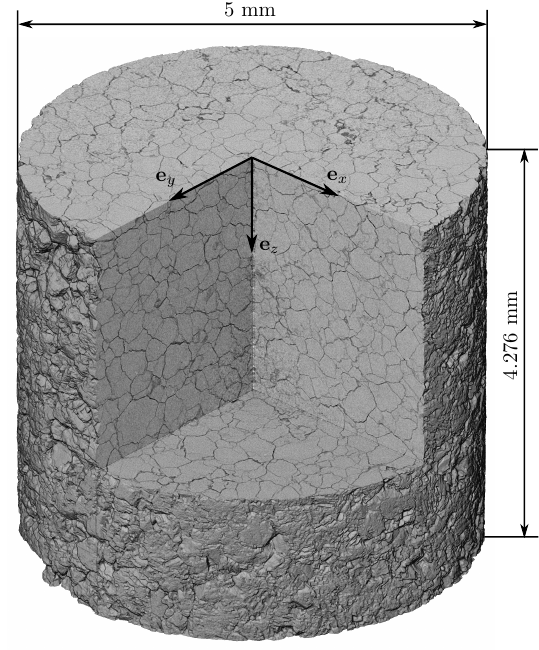}
    \caption{Illustration of the underlying raw \textmu{XRCT} dataset of a thermally treated Bianco Carrara marble sample taken from \cite{Ruf2020}. The dataset along with all meta data can be found in \cite{Ruf2020}, https://doi.org/10.18419/darus-682; licensed under a Creative Commons Attribution (CC BY) license.}
    \label{fig:3D_dataset}
\end{figure}

The scanning of the extracted sample was performed in a self-built, modular, cone-beam {\textmu}-XRCT system, thoroughly described in \cite{Ruf2020b} (please refer to Figure \ref{fig:3D_dataset}). The data-set along with all meta data can be found in \cite{Ruf2020}. For more technical details, please refer to  ~\ref{appendix:materials_and_methods}

\subsection{Noise reduction} \label{sec:noise_reduction}

\begin{figure}[]
    \centering
    \includegraphics[width=0.7\textwidth]{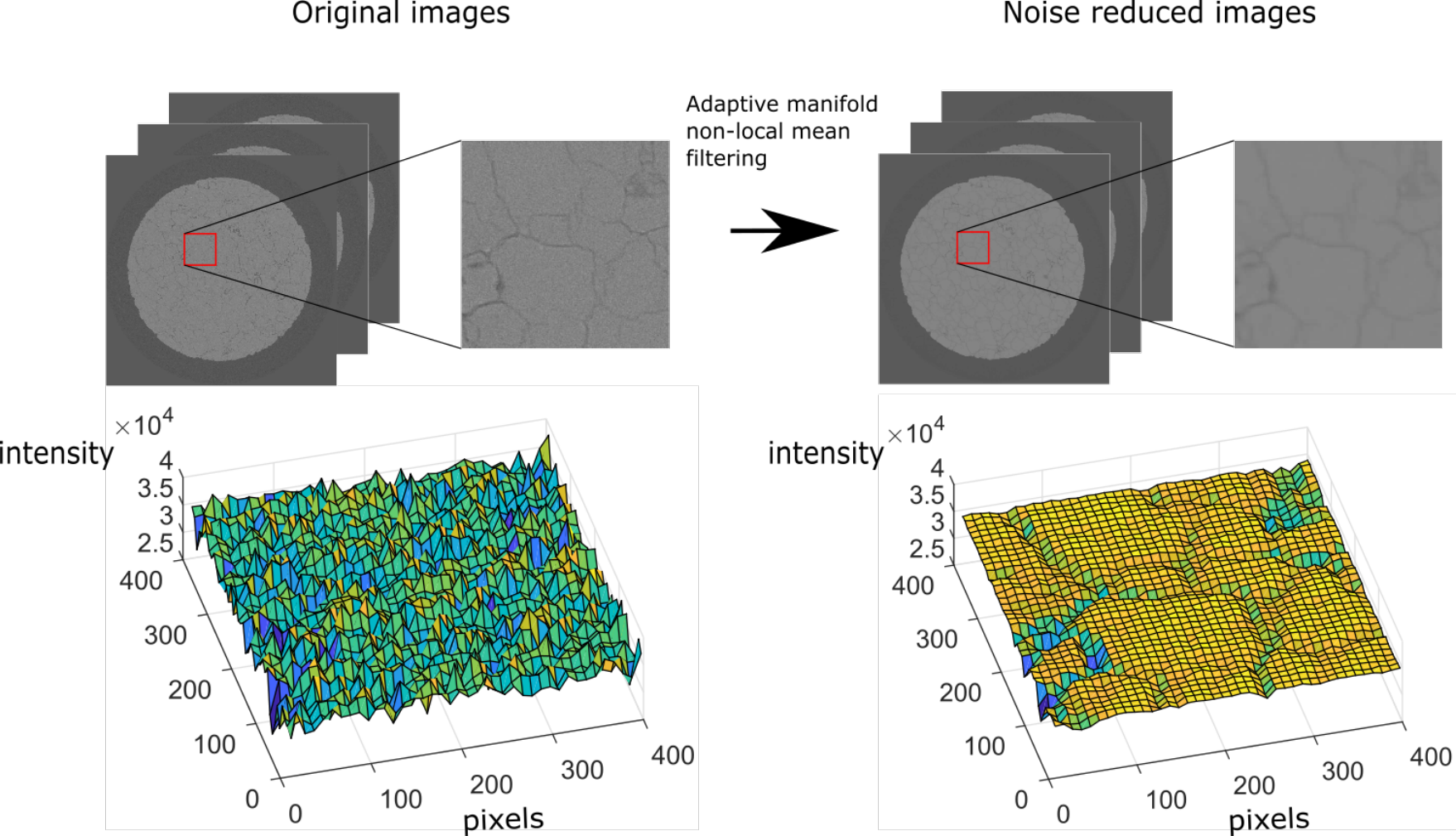}
    \caption{Filtering effects. The images shown above are 2D partial images from the original image (left side) and noise reduced image (right side) by applying 3D adaptive manifolds non-local mean filtering \cite{Gastal2012}. 
    The red box areas are zoomed (400 x 400 pixels). The intensity surfaces for both zoomed images are shown below.}
    \label{fig:noise_reduction}
\end{figure}

In order to conserve the edge information while reducing artefacts, the Adaptive Manifolds Non-Local Mean (AMNLM) method was adopted. As shown in Fig.~\ref{fig:noise_reduction}, this procedure is an essential step to obtain good segmentation results with some of the conventional segmentation methods. The adopted filtering method was applied with the help of the commercial software Avizo\textsuperscript{\textcopyright} (2019.02 ver.). The numerical description of this filtering method in 2D is explained in ~\ref{appendix:noise_reduction}. 

\subsection{Segmentation methods}

\begin{figure}[]
    \centering
    \includegraphics[width =\textwidth]{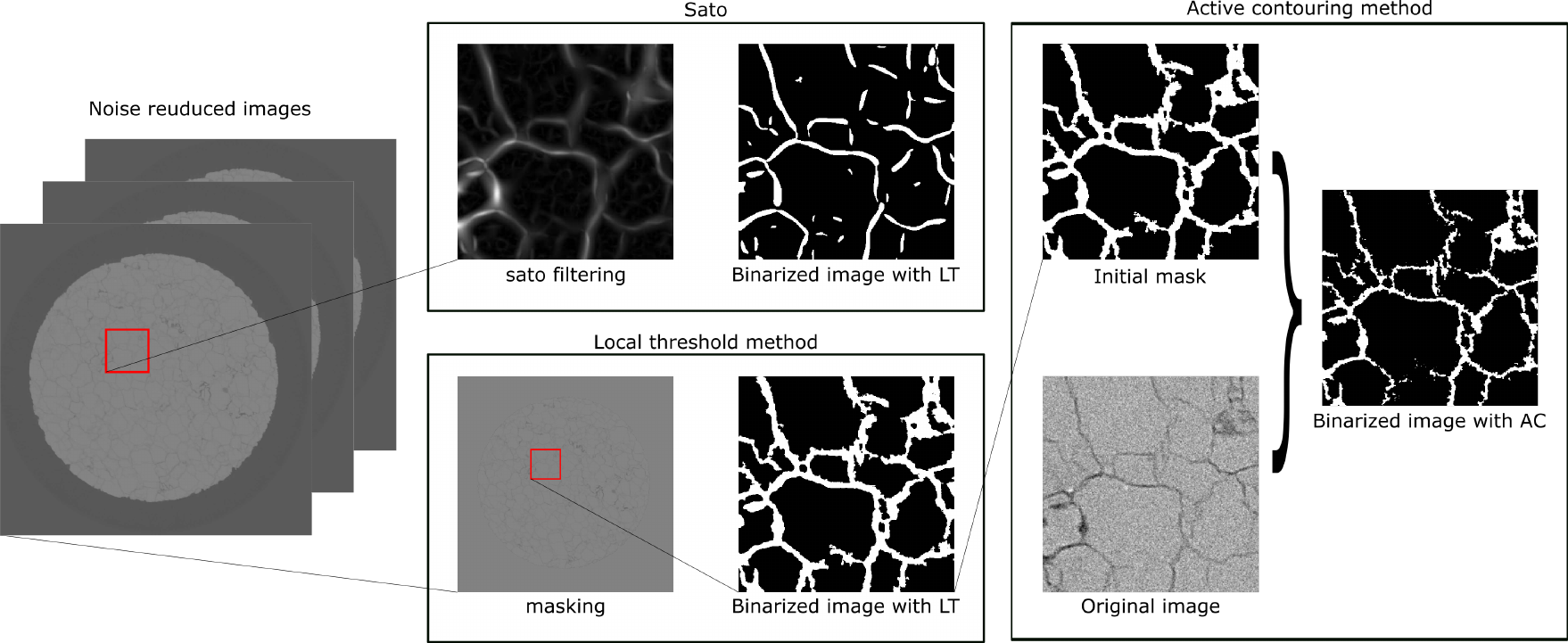}
    \caption{Workflows of the adopted conventional methods: The workflows of segmentation with Sato, Local threshold (LT) method and Active contouring (AC) method are shown with sample images from each step. The magnified region is marked in red.}
    \label{fig:conventional_methods}
\end{figure}

The adopted segmentation methods are introduced and discussed in this chapter. As conventional schemes, the Local threshold, Sato, and Active contouring methods were chosen. Note that the adopted conventional segmentation methods had to be employed with noise-reduced images and also required certain workflows (cf.~ Fig.~\ref{fig:conventional_methods}).

\subsubsection{Local Threshold}\label{sec:local_threshold}
The Local Threshold method is a binarization method, also known as the adaptive threshold method. With this method one can create a threshold map which has the same size as this of the input image for binarization, corresponding to the intensities within the local region. The output of this procedure is a map which includes individual thresholds (local threshold) for each pixel. From the obtained threshold map, binarization is performed by setting the pixels which contain higher intensities than the corresponding location at the threshold map as logical truth, and the others to be logical false. 

For our application, this method was applied on noise-reduced data in order to minimize any artefacts. Smoothing was performed in order to obtain a threshold map with the help of Gaussian filtering Eq. ~\ref{eq:gaussian_filter}. 

Parameters like the offset and the radius of the adjacent area were empirically selected. This procedure was performed with the \texttt{threshold\_local} function from the Python \textit{skimage} library. Small artefacts which remained were removed by employing the \texttt{remove\_small\_object} function from the same library. In order to cope with the modus operandi of the method, before applying the method the outer part of the sample was filled with a mean intensity value extracted from a region of the image where a fracture was not contained.

\subsubsection{Sato filtering}\label{sec:Sato_filtering}

The Sato filtering method is one of the most well-known filtering methods. The potential of this filter in finding structural information is well-matching for the segmentation of a fracture, which often has a long and thin string-like shape in two-dimensional images. An elaborate analysis of the fundamentals of the filtering method can be found in the appendix~\ref{appendix:Sato_filtering}.

\subsubsection{Active contouring}\label{sec:chan-vese}

In image processing, the active contouring method is one of the conventional methods applied in order to detect, smoothen, and clear the boundaries of the features of interest \cite{Caselles1993}. Instead of the conventional active contouring approach, which requires distinguishable gradient information within an image, we adopted the Chan-vese method \cite{Chan2001} which makes use of the intensity information. This has an advantage especially for XRCT applications, as in this case, due to the low signal-to-noise ratio data. An elaborate analysis of the fundamentals of the filtering method can be found in the appendix ~\ref{appendix:Active_contouring}.

\subsubsection{The Random Forest scheme}\label{sec:Random_forest}

The Random Forest method \cite{Amit1997} is using statistical results from multiple classifiers, commonly referred to as decision trees. A decision tree, which is a single classifier, is made of decision nodes and branches which are bifurcated from the nodes. In image segmentation, each node represents an applied filter, such as Gaussian, Hessian, etc., in order to create branches which represent the responses of the filter. These applied filters are also called feature maps, or layers \cite{ArgandaCarreras2017}. In our study, the chosen feature maps, in order to create decision trees, were edge detectors (Sobel filtering, Gaussian difference and Hessian filtering), noise reduction filters (Gaussian blurring) and membrane detectors (membrane projections). The model was trained with the Trainable Weka platform \cite{ArgandaCarreras2017}. The first slice of scanned data at the top side of the sample was used for training with the corresponding result obtained from the active contouring method (cf.~\ref{sec:chan-vese}). On the trained model, diced data \ref{sec:Splitting} was applied and its corresponding predictions were merged as explained in \ref{sec:Merging}.

\subsubsection{The U-net model}\label{sec:U-net}

The U-net model, one of the Convolutional Neural Network (CNN) architectures, proposed by \cite{Ronneberger2015} was applied to segment the fractures. This model finds optimized predictions between input and target by supervising the model with a ground truth (GT). The GT was chosen from the segmentation results from the active contouring combined method \ref{sec:chan-vese} while the input data was raw images which were not treated with any noise reduction techniques.

\begin{figure}[]
    \centering
    \includegraphics[width = \linewidth]{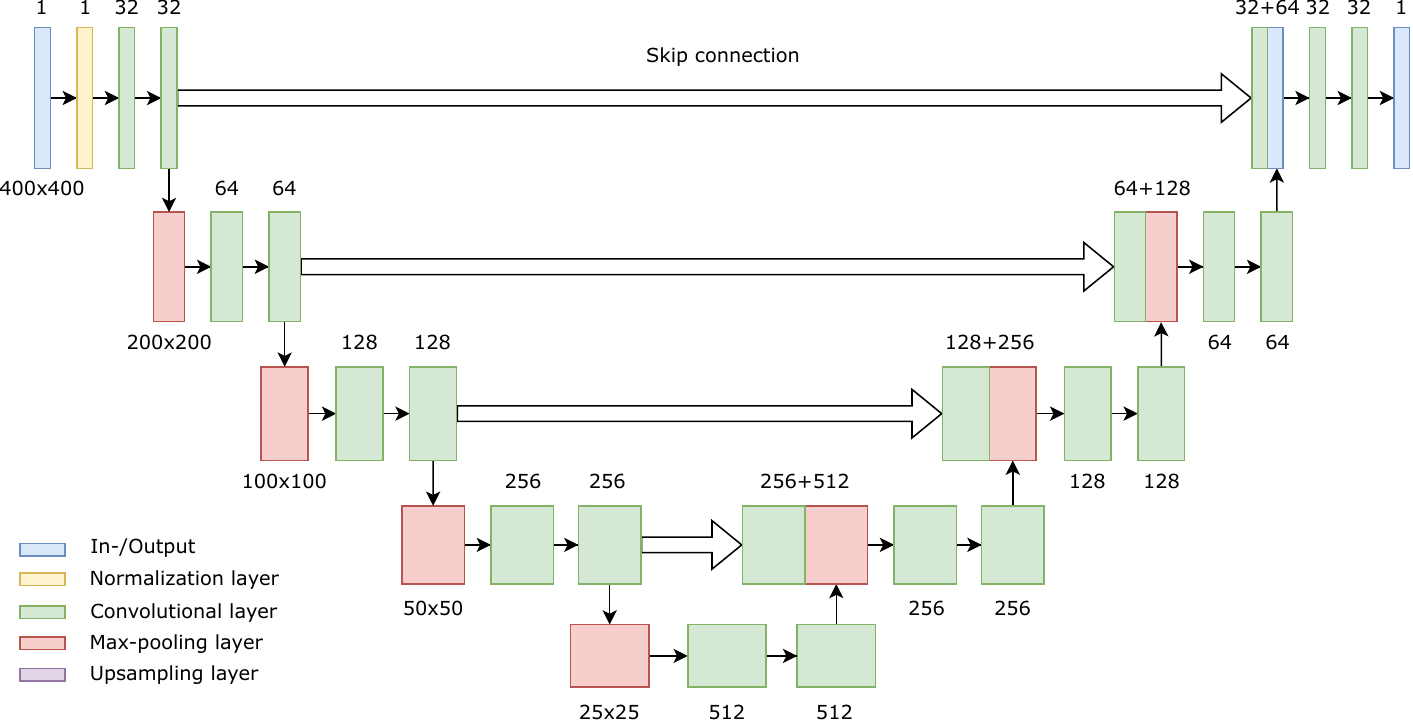}
    \caption{The schematic of the used 2D U-net model}
    \label{fig:2DU-net_model}
\end{figure}

The schematic of the used U-net model is shown in Figure \ref{fig:2DU-net_model}. We adopted a $2\times2$ kernel for max-pooling and the de-convolutional layers. The kernel size of the convolutional layers was decided to be $3\times3$. The adopted activation function for the convolutional layer was the Rectifier Linear Unit (relu) which has a range of output from 0 to $\infty$. It returns 0 for negative inputs and a linear increment up to infinity for positives. On the final output layer, the sigmoid activation function was chosen, which has a non-linear shape and its output range is from 0 to 1. Consequently, the obtained output was a probability map where a number close to 1 was considered as the fracture and the rest was accounted for as non-fracture.

The images which were used for training were cropped into small tiles. This cropping had to be performed due to memory limitations (cf.~\ref{sec:Splitting}). This made each tile to contain varying characteristics which also introduced a difficulty to train a model while taking these different characteristics into account. In order to cope with this problem, the normalization layer, which zero-centers the histogram of the data in the range of -1 to 1, was used previously to insert the data into the model. This was essential to deal with the tiles with ring shape artefacts, which were induced during the reconstruction of the XRCT data. Without this layer the model might not had been able to identify the fractures in such regions due to the different profiles they bear in comparison to the other regions, since the artefacts affect the histogram and shape of the features.

The implementation of the model was done by means of the keras library in Python \cite{chollet2015keras} which includes useful functions for CNN such as convolutional, normalization layers, etc. With the given model details, the input data was separated into small pieces (\textit{Splitting}), trained with GT data (\textit{Training}) and finally it was merged back to the size of the original image (\textit{Merging}). For more information on these processes, please refer to \ref{appendix:machinelearning}.

\section{Results and discussion}
\begin{figure}[]
    \centering
    \includegraphics[width=\linewidth]{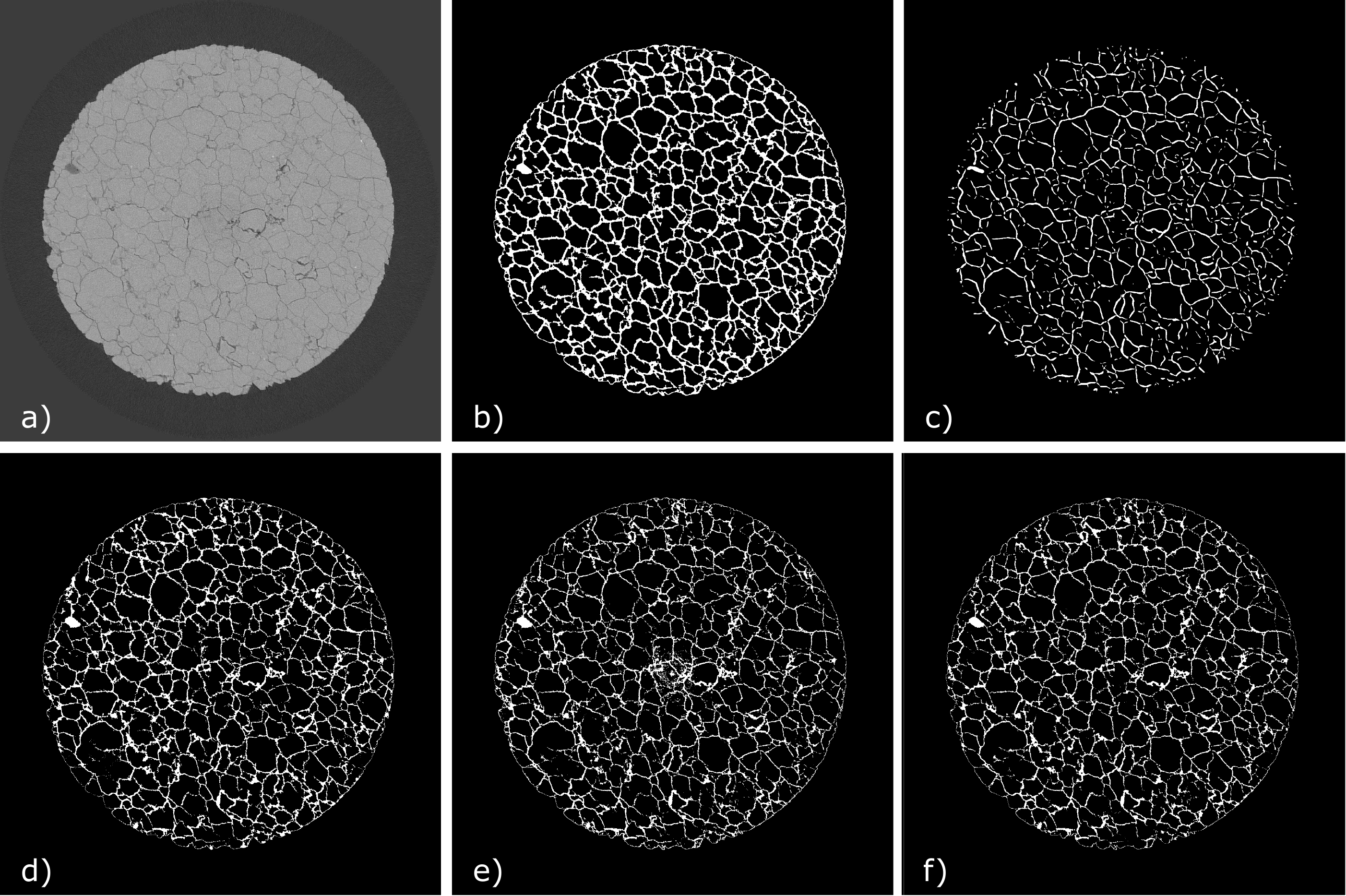}
    \caption{x-y-plane data extracted from the 1000th slice along the z-axis (total 2139). The segmentation results from the a) original data, b) LT, c) Sato, d) Active contouring, e) Random Forest and f) U-net are shown.}
    \label{fig:xy_plain}
\end{figure}

\begin{figure}[]
    \centering
    \includegraphics[width=\linewidth]{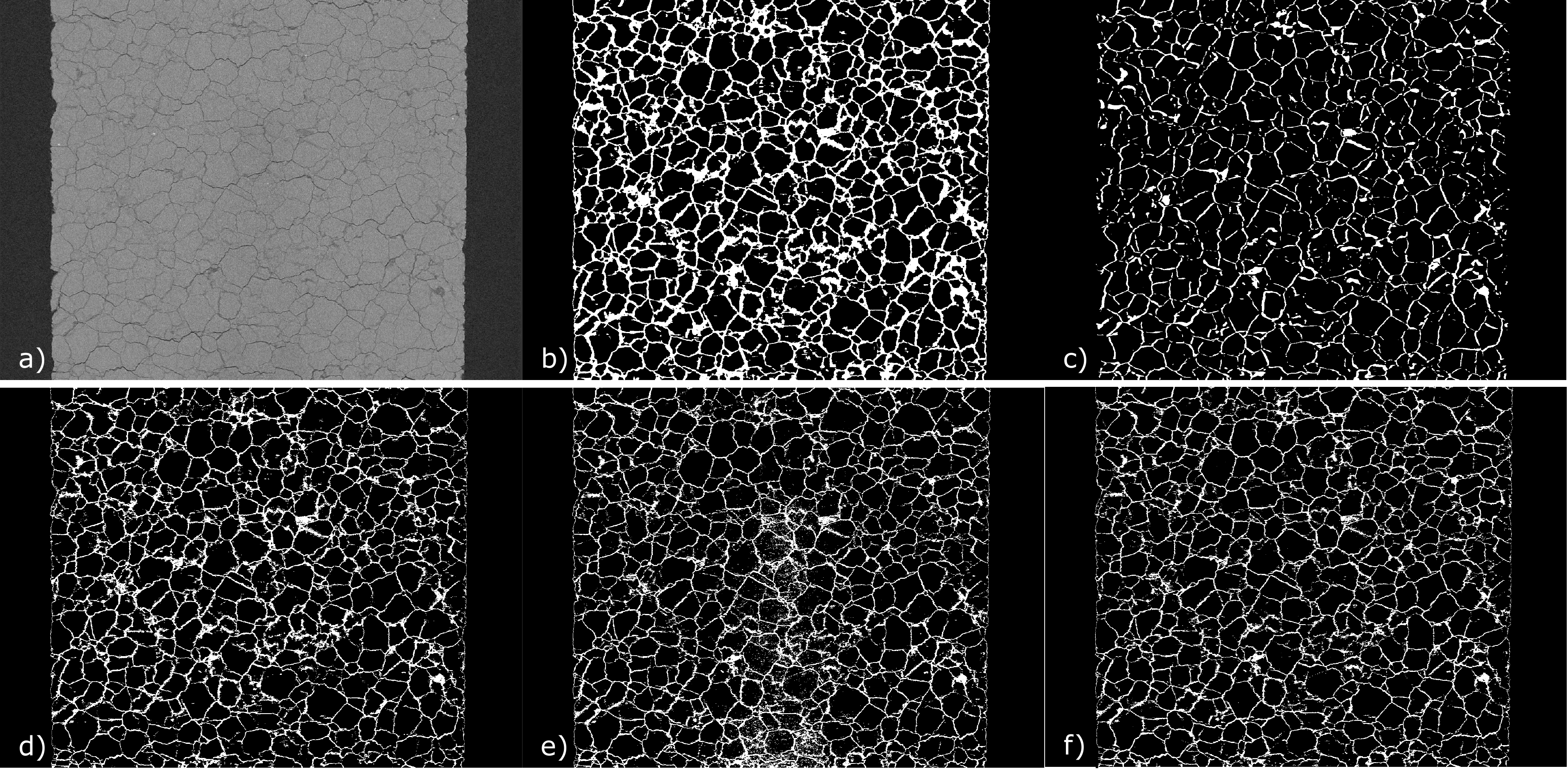}
    \caption{x-z-plane data extracted from the 1400th slice along the y-axis (total 2940). The segmentation results from the a) original data, b) \textbf{LT}, c) Sato, d) Active contouring, e) Random Forest and f) U-net are shown.}
    \label{fig:xz_plain}
\end{figure}
\begin{figure}[]
    \centering
    \includegraphics[width=0.9\linewidth]{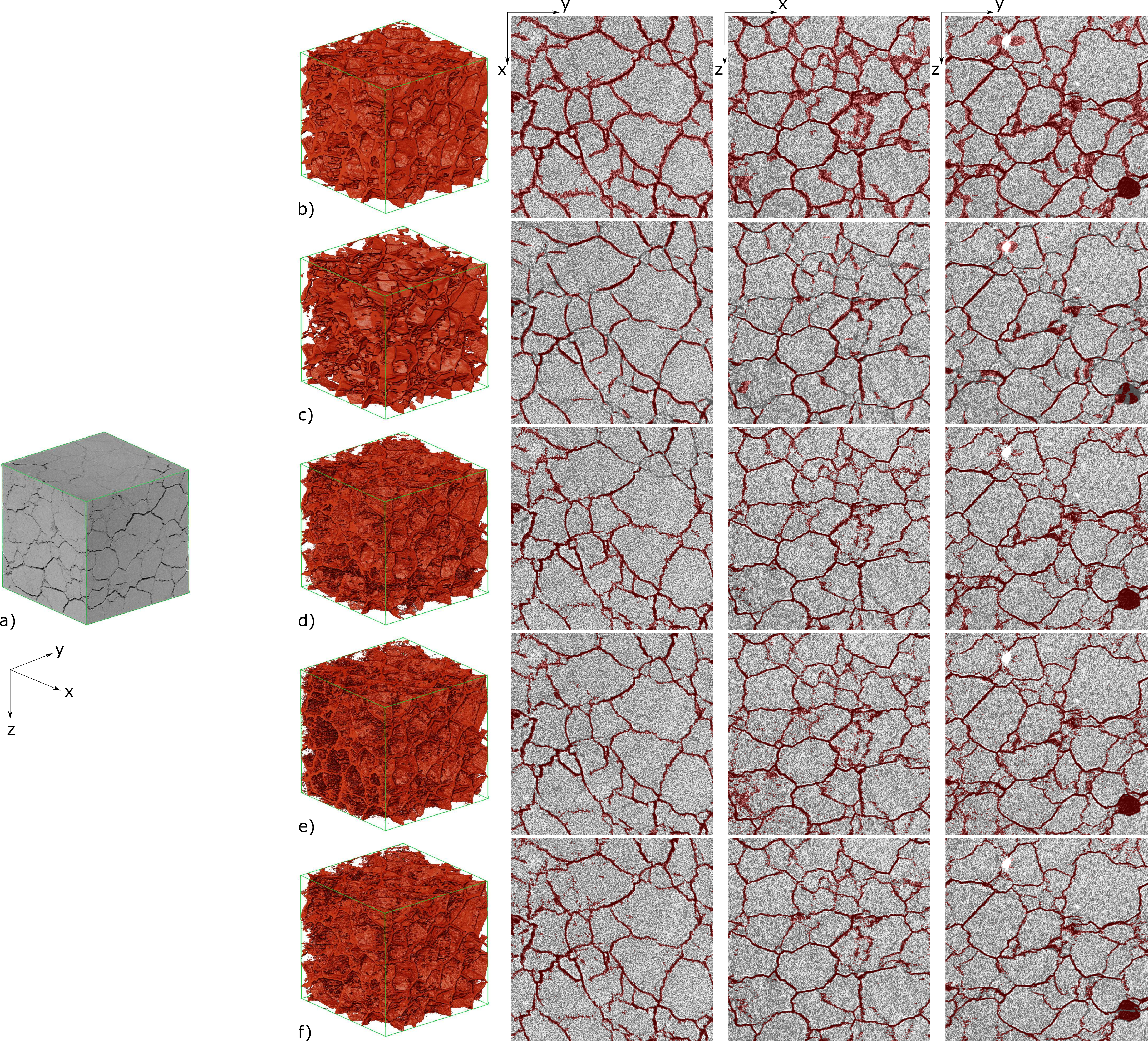}
    \caption{An extracted sub-volume from (1400, 1400, 1000) with a size of $600^3$ voxels is shown in a). The segmentation results from b) Local threshold, c) sato, d) Active contouring, e) Random Forest and f) U-net are shown. The 300th slice in x-y-plane is also shown, as well as the 300th slices in x-z/y-z-plane. Each result is overlayed on the corresponding histogram-enhanced original image for a clear comparison.}
    \label{fig:3D_subvolume}
\end{figure}

We successfully conducted the segmentation of the data acquired by the {$\mu$-XRCT} scanning of a  quenched Carrara marble sample, with all the above-mentioned segmentation schemes. By performing 2D segmentation on each slice of scanned data and stacking the segmentation results, the full 3D segmented fracture network was able to be acquired for the entire scanned domain. 

Parameters like standard deviation, offset, smoothing factors for active contouring etc., of the methods based on conventional schemes, such as the Local threshold~\ref{sec:local_threshold}, Sato~\ref{sec:Sato_filtering} and active contouring~\ref{sec:chan-vese}, were fine-tuned empirically. It was time consuming to decide these parameters, since we had to find the appropriate parameters which would give the best results along the entire data-stack, while some slices contained different histogram profiles, or artefacts. Since the conventional schemes showed a tendency to be very sensitive to artefacts, and the corresponding outputs varied significantly with the applied parameters, applying noise reduction (\ref{sec:noise_reduction}) with carefully chosen parameters was a fundamental step before applying any conventional method, in order to minimize the error and obtain good results. In the case of machine learning-based segmentation, the applied methods had the ability to classify the fracture networks within the original data. Since these machine learning-based algorithms do not require any noise reduction, this was one of the biggest advantages which not only reduced the computation time, but also reduced the input parameters tuning time.

In Figure~\ref{fig:xy_plain}, the segmentation results of the top-viewed specimen (x-y-planar size of (2940 $\times$ 2940 pixels) are shown. We were able to acquire sufficiently good segmented results for the whole x-y-plane despite the beam hardening effect at the edges, and the ring artefacts in the middle of the sample. We selected the 1000th slice of the scanned data on the z-axis, which is located at the middle of the sample. Note that this slice was never used as training data for the machine learning-based methods. In Figure~\ref{fig:xz_plain}, the segmentation results from the side-viewed sample (x-z-planar size of 2940 $\times$ 2139 pixels) are demonstrated. These images were obtained by stacking the 2D segmentation results from x-y-planar images along the z-axis and extracting the 1400th slice on the y-axis.  By showing the results from side and top, we demonstrate that there was no observable discontinuity of the segmented fractures which could be caused by performing 2D segmentation and splitting.

Based on the 2D results in both Figures \ref{fig:xy_plain} and \ref{fig:xz_plain}, we were able to identify some advantages and limitations for each segmentation method. With the local threshold method we were able to obtain the segmented results which included most of the features within the data-set. The method definitely has the benefit of easy implementation and short computation time. Additionally, it was consistent providing a continuous shape for the fracture network. However, the thickness of the segmented fracture was unrealistically large. Also, due to the small contrast of the features, the method faced challenges distinguishing the features precisely when the fractures were located close to each other. In the case of the Sato method there was no significant benefit observed. The method was able to capture some of the fractures which had an elevated contrast however, it was not able to identify the faint fractures which had a lower contrast. In addition, the method also showed a limitation to classify the fractures at the points where multiple fractures were bifurcating. This was expected as a characteristic feature of a Hessian matrix-based method. Additionally, the pores detected by this method tended to appear narrower than their real size. Please note that the outer edges between the exterior and inner part of the specimen were excluded with this method. 

With the help of the active contouring method we were able to obtain better results than the other conventional segmentation methods. The fractures were detected with thinner profiles, and pores were also identified accordingly, despite of the ring artefacts at the central area of the image (as shown in Figure~\ref{fig:xy_plain}). However, the fractures close to the outer rim still appeared thicker. This was caused by the beam hardening effect, which induces spatially-varying brightness. Additionally, some parts which contained faint fractures were recognized with a weak connection, which meant that the segmented fractures were not well identified. This was induced due to a variation of contrast of the fractures within the region of interest. Although we cropped the images into small tiles before applying the method, the fractures with low contrast tended to be ignored when the fractures of larger contrast were mostly taken into account. Despite the observed drawbacks, the output of the method gave us sufficiently well segmented fractures thus, we adopted this result as a "truth" image for the machine learning-based models. 

In the case of the Random Forest method, where training happens with the use of a single slice of output of the active contouring method, the trained model was able to detect the fractures with finer profiles, better than any other conventional method. More specifically, it was able to detect well the fractures when two fractures were located close to each other, which were identified as a merged one with the conventional methods. However, as shown in Figures \ref{fig:xy_plain} and \ref{fig:xz_plain}, it seems that the method is sensitive to ring artefacts based on a significant number of fuzzy voxels classified as fractures at the central part of the image, while this was not the case.

The segmentation results of the U-net method  outperformed any other method employed in this comparison. With the help of the trained model, we were able to obtain clear segmentation results despite beam hardening and ring artefacts. In Figure~\ref{fig:xy_plain}, all the fractures which were of high and low contrast were efficiently classified while maintaining their aperture scale. In addition, with the help of the overlapping scheme, we obtained continuous fractures even though we trained the model and computed the predictions with cropped images.


For the sake of detailed visualization, we demonstrate the segmentation results in a 3D sub-volume (600$^3$ voxels) in Figure~\ref{fig:3D_subvolume} with overlaid results on the contrast-enhanced original image on x-y- and x-z-planes at the middle (300th) of the sub-volume. The contrast of images was improved by using the histogram equalization method which adjusted the intensities of the image by redistributing those within a certain range \cite{Acharya2005}. In the 3D visualized structure with each of the adopted methods in Figure~\ref{fig:3D_subvolume}, we were able to obtain an overview of their unique responses. The segmented fractures from each method are marked in red. In the case of the local threshold method, as mentioned earlier, the method was able to recognize all the features however, the fractures were identified with a broader size compared to those in the original data. In addition, the outlines of the results were captured with rough surface profiles. Although the segmentation result with the method showed sufficiently well-matched shapes in general, from the middle of the x-z-plane overlaid image we could conclude that the method was not so appropriate to detect sophisticates profiles, especially when the fractures were next to each other since these features were identified as a cavity. In the case of the Sato method, as we showed in Figure~\ref{fig:xy_plain} and Figure~\ref{fig:xz_plain}, we were able to conclude from the 3D visualized data that the method was not so efficient to detect the fractures. The roughness of the surface was not able to be accounted for, and an unrealistic disconnection between the fractures was observed due to unidentified fractures with this method.

In the 3D visualized result with the active contouring method in Figure~\ref{fig:3D_subvolume}, the fractures were generally well classified and visualized. However, as it is shown in the overlaid images in the right top corner, some fractures were not detected due to their low contrast. In addition, the outlines of the fractures with sophisticated shapes were not able to be well-classified.

Although we show in Figures \ref{fig:xy_plain} and \ref{fig:xz_plain} that the result with the Random Forest method had challenges dealing with ring artefacts, the result in general showed accurate fracture profiles as it is demonstrated in Figure~\ref{fig:3D_subvolume} with an overlaid x-y-plane image. The segmentation results showed that the method was able to detect fractures with both high and low contrast. However, in the overlaid x-z-plane image, we can observe that the left bottom side contained a lot of of defects which were also found in the 3D visualization from the rough surface in the left corner. Thus, we were able to conclude that the segmentation was not well performed at some locations.

In the case of the U-net model, the segmented fractures were well matched with the fractures in the original data. Consequently, a clear outlook of fractures was also obtained in the 3D visualized data. 

Based on mentioned environment in \ref{appendix:materials_and_methods}, the noise reduction took 8 hours for the entire data-set. The total computation time to obtain the segmentation results were 1.3 hours with local threshold, 35.6 hours with Sato, 118.2 hours with active contouring, 433.7 hours with Random Forest and 4.1 hours with the U-net method. Note that the merging and cropping steps which were necessary in order to employ the active contouring, Random Forest and U-net method are included in this time estimate (approximately 40 min). The training times for the machine learning based methods took 3.5 hours for the Random Forest and 21 hours for the U-net method.

\begin{table*}[t]
\caption{Summary of the estimated porosity for all adopted segmentation methods in comparison to experimentally determined value}
\begin{tabular}{l c r}

\tophline
Methods & Estimated porosity ($\%$) & Processing time (hours)\\ 
\middlehline
Local Threshold& 25 & 1.3\\
 Sato& 12  & 35.6\\
 Chan-vese&15 & 118.2\\
 Random Forest&16 & 433.7\\
 U-net & 14 & 4.1\\
 Experimental measurement & 3 & -\\
\bottomhline
\end{tabular}\label{tab:summary_of_porosity}
\belowtable{} 
\end{table*}

By counting the number of fracture-classified voxels and dividing with the total number of voxels, we were able to compute an approximated porosity out of the images. The summary of the estimated porosity from all segmentation methods is shown in Table \ref{tab:summary_of_porosity}. In this estimation, the outer rim was excluded in the counting by applying a mask which was obtained by the combination of the Otsu method for binarization, and the fill and erosion method \cite{Dougherty1994}. With the help of this evaluation, we acquired the porosity of the local threshold method: 25 \%, Sato method: 12 \%, active contouring method: 15 \%, Random Forest method: 16 \% and U-net method: 14 \%. These values were significantly larger than the porosity obtained from measurements of the sample which was around 3 \%. The porosity of the thermally treated sample was obtained from measurements of volume changes before/after quenching. 

This mismatch between the numerically estimated values and the measurement for porosity was expected. The resolution of the scanned data was not high enough to accurately estimate the size of a single fracture aperture, although we adopted the highest resolution we could achieve \cite{Ruf2020}, which was this of 2 microns/voxel, since this resolution is comparable to the aperture of a fracture. In addition, our intention was to obtain geometrical information of fracture networks from a wide field of view. Nevertheless, we used these results to speculate about the comparative accuracy of the segmentation techniques, without emphasising on the accurate prediction of porosity, at this stage. 

As we showed in Figure~\ref{fig:3D_subvolume}, the apertures in the result of the local threshold method were significantly overestimated. In the case of the Random Forest and active contouring method, although the methods were able to recognize fractures with thin profiles, the porosity was exaggerated due to poorly classified pixels. For the porosity estimation based on the Sato method, which showed the lowest value, we demonstrated that there were many undetected features. By considering these facts, the result from the U-net method again gave us the best result among the compared methods.

\conclusions  

In this work, we demonstrate the segmentation workflows for three different conventional segmentation methods, and two machine learning methods for $\mu$-XRCT data from a dry fracture network which was induced by thermal quenching in Carrara marble. Despite the low contrast of the fractures in the available data, due to their aperture being very close to the resolution of the $\mu$-XRCT, we were able to successfully segment 3D fracture networks with the proposed segmentation workflows.

With the acquired segmentation results we conducted a quality and efficiency comparison and showed that the results with the U-net method were the most efficient and accurate ones. Among the applied conventional segmentation schemes we were able to obtain the best quality of results with the active contouring method, and the best time efficiency with the local threshold method. For the adopted machine learning schemes, we trained the models with results from the active contouring method. This was advantageous compared to manual labelling, especially since manual labelling was far from being the best option given that the fractures were of low contrast. Manual labelling work would be too arduous to obtain sufficient and accurate training data. In addition, we also showed that the defects which were detected with the conventional methods were improved with the trained machine learning methods. 

In order to perform the segmentation with conventional methods, the application of the adaptive manifold non-local mean filter, which is one of the noise reduction techniques, was a fundamental step due to the inherent noise of the scanned data. However, with the machine learning methods this step was not necessary. We were able to obtain fully segmented data by providing less filtered data to the training model. This was quite advantageous compared to the conventional approaches since the computation time and resources requirements were significantly reduced. Additionally, the fine-tuning of the input parameters, which were mandatory for the conventional methods, did not play a significant role in the attempt to get good segmentation results.

We showed that obtaining a full 3D structure of segmented data can be performed efficiently by employing the proposed segmentation workflows in 2D images. Thus, the big volume of data could be dealt with a conventional workstation without requiring any advanced properties of a specific image-analysis workstation.
Although, we showed that the resolution of the data to estimate the exact size of apertures of fracture networks was not sufficient, by comparing with the actual measurement of porosity, the geometrical information was acquired without demanding any additional procedure. For future work, we intend to correlate the acquired geometrical information regarding fractures with measurement results obtained from (mercury) porosimetry. In addition, characterization of fractures in terms of lengths and angle will be further investigated.


\codeavailability{The first version of the \textit{fracture network segmentation} used for classification of sophisticated shapes of fracture network from quenched carrara marble data \cite{Ruf2020b} is preserved at doi:10.18419/darus-1847, available via no registration and create commons attribution conditions, and developed openly at python (https://www.python.org/), matlab (https://www.mathworks.com) and beanshell (https://beanshell.github.io/).} 

\dataavailability{The tomographic data of quenched carrara marble \cite{Ruf2020b} used for validation efficiency and testing effectiveness of adopted segmentation methods of in the study are available at DARUS, \textit{micro-XRCT data set of Carrara marble with artificially created crack network: fast cooling down from 600°C} via doi:10.18419/darus-682 with no registration and creative commons attribution conditions.} 




\appendix
\counterwithin*{equation}{section}
\renewcommand{\theequation}{A.\arabic{equation}}
\section{Technical specifications}\label{appendix:materials_and_methods}
An open micro-focus tube \emph{FineTec FORE 180.01C TT} with a tungsten transmission target from Finetec Technologies GmbH, Germany, in combination with a \emph{Shad-o-Box~6K~HS} detector with a CsI scintillator option from Teledyne DALSA Inc., Waterloo, Ontario, Canada with a resolution of \SI[product-units = single]{2940}x{ 2304} {pixels} and a pixel pitch of \SI{49.5}{\micro\meter} and 14-bit depth resolution was employed. For more details about the system please refer to \cite{Ruf2020}. The highest achievable spatial resolution with this system is about 50 line-pairs/mm at \SI{10}{\%} of the Modulation Transfer Function (MTF), which is equal to a resolvable feature size of about \SI{10}{\micro\meter}. To achieve this, we are using a geometric magnification of 24.76 which results in a final voxel size of \SI{2.0}{\micro\meter}. With these settings the Field Of View (FOV) is about \SI{5.88}{\milli\meter} in the horizontal direction and \SI{4.61}{\milli\meter} in the vertical one. Consequently, the sample could be scanned over the entire diameter of \SI{5}{\milli\meter}. The overall physical size of the FOV and consequently the 3D volume is required for the definition of an appropriate Representative Elementary Volume (REV) for subsequent Digital Rock Physics (DRP), which is at the same time the challenge between a sufficient large FOV and a high spatial resolution to resolve the fractures. For further details about the image acquisition settings cf. \cite{Ruf2020b} and \cite{Ruf2020}.  The reconstruction was performed using the Filtered Back Projection (FBP) method implemented in the software \emph{Octopus Reconstruction (Version 8.9.4-64 bit)}. The reconstructed 3D volume consists of \SI[product-units = single]{2940}x{2940}x{2139} voxels, saved as an image stack of 2139 16-bit *.tif image files of \SI[product-units = single]{2940}x{2940} resolution. Noise reduction was deliberately omitted in the reconstruction process in order to provide an interested reader with the possibility to use their own adequate noise filtering methods. The 3D data-set with an extraction to better show the homogeneous fracture network is illustrated in Figure~\ref{fig:3D_dataset}. The bright gray area represents the calcite phase whereas the dark gray area in between the calcite phase shows the initiated micro-fractures which were generated by the thermal treatment. As a consequence of the thermal treatment, a bulk volume increase of about \SI{3.0}{\%} under ambient conditions can be recorded if a perfect cylindrical sample shape is assumed. The noise reduction and segmentation workflows were performed with the hardware specification of intel(R) Core(TM) i7-8750H CPU @2.2GHz, 64GB of RAM and Nvidia Quadro P1000 (4GB).

\section{Noise reduction schemes}\label{appendix:noise_reduction}
\renewcommand{\theequation}{B.\arabic{equation}}
The AMNLM filter operates in two parts: 1.) Creation of matrices (manifolds) which are filled with weighted intensities by coefficients. 2.) Iterative processing on created matrices. In the first part, in order to reduce the complexity and increase the efficiency of the filter, \cite{Gastal2012} applied the Principal Component Analysis (PCA) \cite{pearson1901liii} scheme on the created matrices. In the second part, sequential steps were presented to adjust intensities of pixels based on components of created and PCA-treated matrices. These procedures are called \textit{Splatting}, \textit{Blurring} and \textit{Slicing}. The \textit{Splatting} and \textit{Blurring} are iterative steps on components of created matrices with numerical modifications (adaptive manifolds). After these iterative procedures, \textit{Slicing} is performed which is merely summing the outputs from the iteration and normalizing it.

From the creation of manifolds the coefficients to weight intensities were designed to have a shape of a Gaussian distribution in the kernel. A typical two-dimensional isotropic Gaussian kernel can be derived as follows,
\begin{linenomath*}
\begin{equation} \label{eq:Gaussian}
     G((x, y);\sigma_f) = A\exp{\bigg(-\frac{x^2 + y^2}{2\sigma_f^2}\bigg)}
\end{equation}
\end{linenomath*}
where $A$ is an arbitrary chosen amplitude of the Gaussian profile, $x$ and $y$ are input vectors which have a mean value of zero, and $\sigma_f$ is the standard deviation of Gaussian distribution.

By using the above profile and input vectors, a finite sized Gaussian kernel $G_k$ could be defined. With the created Gaussian kernel, the manifolds of weighted intensities $f_A$ could be created as follows,
\begin{linenomath*}
\begin{equation}
    f_A = vec(G_k)^T\cdot vec(I)
\end{equation}
\end{linenomath*}
where $vec(\cdot)$ is a vectorization operator which transforms a matrix into a column vector, and $I$ is the 2D image matrix which contains intensities.

Each dimension of $f_A$ contains weighted intensities from the image. The dimensionality of the PCA scheme was reduced in order to boost the computation efficiency while minimizing the loss of information \cite{Jollife2016}. This could be achieved by finding the principal components which would maximize the variance of data in dimensions of $f_A$. The principal components could be obtained by computing eigenvalues and theirs corresponding eigenvectors of covariance matrix $C$. The covariance matrix $C$ of $f_A$ is acquired as follows,
\begin{linenomath*}
\begin{equation} \label{eq:covariance}
    C = (f_A - \mathcal{E}(f_A))^T (f_A - \mathcal{E}(f_A))
\end{equation}
\end{linenomath*}
where $\mathcal{E}$ operator is to compute means of its arguments.

Among the eigenvalues of the obtained $C$, the bigger ones indicate higher variances of the data in the corresponding eigenvector space \cite{Jollife2016}. Thus, after sorting the corresponding eigenvectors in an order of magnitudes of eigenvalues, by performing multiplication with these eigenvectors and $f_A$, the manifolds $f_A$ could be ordered in order of significance. Based on this, the user defined number of dimensionality $s$ is adopted and the manifold $f_A$ is recast into $f_S$ by extracting $s$ number of principal components.

\paragraph{Splatting}
With the given extracted manifolds $f_S$, the Gaussian distance-weighted projection is performed as follows,
\begin{linenomath*}
\begin{equation}
    \Psi_0 = \exp{\bigg(-\frac{\sum_{i=1}^S |f_{i} - \eta_{i}|^2}{2\sigma_r^2}\bigg)},
\end{equation}
\end{linenomath*}
with $\Psi_p$ being the coefficients to weight the original image (projection), $\sigma_r$ being a chosen filtering parameter which states the standard deviation of filter range, and $\eta_S$ denoting the \textit{Adaptive manifolds} which are noise reduced responses by a low-pass filter(eq.~\ref{eq:low_pass}) at the first iteration. These adaptive manifolds will be further updated via an iteration process. Consequently, the Euclidean distances between each of $f_S$ and $\eta_S$ are computed and integrated into $\Psi_0$ by considering the standard deviation $\sigma_r$.

In order to obtain {\textit{Adaptive manifolds}}, the numerical description of the low-pass filter for a 1-D signal in the work of \cite{Gastal2012} is as follows,
\begin{linenomath*}
\begin{equation} \label{eq:low_pass}
    S_{out}[i] = S_{in}[i] + \exp{\bigg(-\frac{\sqrt{2}}{\sigma_s}\bigg)}(S_{in}[i-1] - S_{in}[i]),
\end{equation}
\end{linenomath*}
where $S_{out}$ is the response of the filter, $S_{in}$ is the signal input, and $\sigma_s$ is a filtering parameter which states the spatial standard deviation of the filter. $i$ denotes the location of a pixel in the image. Note that this low-pass filter has to be applied to each direction in an image due to its non-symmetric response corresponding to the applied direction. With this relation, the collected response has a smoother profile than the original signal. 

By performing element-wise multiplication on the acquired $\Psi_p$ and the original image $I$, the weighted image $\Psi$ could be obtained such as,
\begin{linenomath*}
\begin{equation}
    \Psi = \Psi_0 \circ I.
\end{equation}
\end{linenomath*}
where $\circ$ denotes the Schur product \cite{Davis1962}.

\paragraph{Blurring}
In the computed weighted image $\Psi$ and the \textit{adaptive manifolds} $\eta_S$, Recursive Filtering (RF) \cite{Gastal2011} is applied so that the values of each manifold in $\eta_S$ can be blurred accordingly in $\Psi$ and $\Psi_0$. \cite{Gastal2011} proposed to conduct this procedure in a down-scaled domain so that further smoothing could be expected after the following up-scaling interpolation. The down-scaling factor $d_f$ is calculated as such (e.g., $d_f = 2$ states a half-scaled domain),
\begin{linenomath*}
\begin{equation}
    d_f = max(1,2\,{\lfloor}\log_{2}(min(\sigma_s/4, 256\,\sigma_r)){\rfloor})
\end{equation}
\end{linenomath*}
by taking into account the adopted spatial/intensity filter range of standard deviations $\sigma_s$ and $\sigma_r$.

\paragraph{Slicing} 
As mentioned above, the \textit{Splatting} and \textit{Blurring} steps  are performed in an iterative manner and, at each iteration step, a weight projection matrix $\Psi_0$ and a weighted image $\Psi$ are created. By defining the \textit{Blurring} responses of these with RF as ${\Psi}^{blur}(k)$ and ${\Psi_0}^{blur}(k)$ where $k$ represents iteration number, the final normalized result of the filter $I_f$ is as follows,
\begin{linenomath*}
\begin{equation}\label{eq:final_response}
   I_f = \frac{\sum_{k=1}^{K}{\Psi}^{blur}(k)\circ{\Psi_0(k)}}{\sum_{k=1}^{K}{\Psi_0}^{blur}(k)\circ{{\Psi_0}(k)}}
\end{equation}
\end{linenomath*}
where $K$ is the total numbers of created adaptive manifolds.
Note that up-scaling is performed by bilinear interpolation on every $\Psi^{blur}$ and ${\Psi_0}^{blur}$ to recover the domain size from the down-scaled one due to \textit{Blurring} step.

\paragraph{Adaptive manifolds} 
In the first iteration, the $\eta_S$ are obtained from the low-pass filter response of $f_S$ with the use of eq.~\ref{eq:low_pass}. The responses of the low-pass filter locally represent the mean value of the local intensities, which reflects well the majority of intensities within the sub-domain while it lacks variation. By taking into account this drawback, \cite{Gastal2011} introduced a hierarchical structure to include better representative information from additionally created dimensions.  These additional dimensions are created by clustering the pixels of $f_S$. By taking each of $\eta_S$ as default points, the pixels which had  intensities above $\eta_S$ and below are classified accordingly.

This type of clustering, i.e. finding which pixels are located above and below $\eta_S$, could be acquired by the eigenvalues of the covariance matrix of $(f_S - \eta_S)$ with eq.~\ref{eq:covariance}. From the dot product of the obtained eigenvalues and $(f_S - \eta_S)$, the clustered pixels are,
\begin{linenomath*}
\begin{equation}
\begin{cases}
        p_i \in C_{-}, & \text{if J < 0},\\
        p_i \in C_{+}, & \text{otherwise},
\end{cases}
\end{equation}
\end{linenomath*}
where $J$ is the dot product, $p_i$ is pixel i, $C_{-}$ and $C_{+}$ denote the clusters with pixels below and above the manifold, respectively.

Based on the computed clusters $C_{-}$ and $C_{+}$, \textit{Adaptive manifolds} are calculated as follows, 
\begin{linenomath*}
\begin{equation}
    \eta_{S-} = \mathcal{L}_p(C_{-}\circ(1-\Psi_0)\circ{f_S},\sigma_s)\oslash \mathcal{L}_p (C_{-}\circ{(1-\Psi_0)},\sigma_s),
\end{equation}
\end{linenomath*}
and 
\begin{linenomath*}
\begin{equation}
    \eta_{S+} = \mathcal{L}_p(C_{+}\circ(1-\Psi_0)\circ{f_S},\sigma_s)\oslash \mathcal{L}_p (C_{+}\circ{(1-\Psi_0)},\sigma_s),
\end{equation}
\end{linenomath*}
where $\mathcal{L}_p$ indicates the low-pass filter in eq.~\ref{eq:low_pass} and $\oslash$ is an element-wise division operator \cite{Wetzstein2012}. Note that the low-pass filter is performed on down-scaled domain such as the step in \textit{Blurring}. Thus, in order to match the size of matrix, $\eta_S$ is up-scaled before it is applied to \textit{Splatting} step. These steps are performed iteratively (compare eq.~\ref{eq:final_response}). Thus, $K = 2^h-1$ numbers of manifold are created while $h$ is defined as follows,
\begin{linenomath*}
\begin{equation}
    h = 2+max(2, \lceil(\lfloor\log_2\sigma_s\rfloor-1)(1-\sigma_r)\rceil).
\end{equation}
\end{linenomath*}

\section{Sato filtering}\label{appendix:Sato_filtering}
\renewcommand{\theequation}{C.\arabic{equation}}
The filter consists of two fundamental parts: a Gaussian filter and a Hessian matrix. In the Gaussian filtering part, with the Gaussian shape profile in eq.~\ref{eq:Gaussian}, a finite-sized kernel $w$ is designed by defining the input vectors $(x,y)= [-l_w, l_w]$ with a mean value of zero, and $l_w$ being a positive integer defined by the standard deviation $\sigma_f$ and truncate factor $t_f$ e.g. ($l_w = t_f*\sigma_f + 0.5$). For the sake of stability of the kernel behavior, the kernel is normalized by dividing with the sum of all of its elements. Finally, the Gaussian filter can be expressed by means of the kernel with a convolution operator $\circledast$ such as, 
\begin{linenomath*}
\begin{equation}\label{eq:gaussian_filter}
    I_{f}(\mathbf{X}) = G(w(x,y);\sigma_f) \circledast I(\mathbf{X}),
\end{equation}
\end{linenomath*}
where $I$ is an intensity matrix (input image). in the spatial domain $\mathbf{X}$ and $I_{f}$ being Gaussian-filtered image.

The Hessian matrix is a matrix of second order partial derivatives of intensities in the x and y directions, and is often used to trace local curvatures with its eigenvalues. By bringing this concept to an image we can deduce a local shape at selected pixels, while observing the changes of intensity gradients. The numerical description of a Hessian matrix with a Gaussian filtered image at selected pixel positions $z$ is as follows,

\begin{linenomath*} 
\begin{equation}\label{eq:hessian}
    \nabla^2 I_f(z) =  \begin{bmatrix}
    \frac{\partial^2}{\partial{x}^2}I_{f}(z) & \frac{\partial^2}{\partial{x}\partial{y}}I_{f}(z)\\
    \frac{\partial^2}{\partial{y}\partial{x}}I_{f}(z) & \frac{\partial^2}{\partial{y}^2}I_{f}(z)
    \end{bmatrix}
\end{equation}
\end{linenomath*}
With eq.~\ref{eq:gaussian_filter} we can re-write the elements of the matrix as the convolution of the input image and the second partial derivatives of the Gaussian shape function. For example, the first element of the Hessian matrix is,
\begin{linenomath*}
\begin{equation}\label{eq:Hessian_element}
    \frac{\partial^2}{\partial{x}^2}I_f(z) = \frac{\partial^2}{\partial{x}^2}G(w(x,y);\sigma_f)\circledast I(z_w)
\end{equation}
\end{linenomath*}
where $z_w$ is the kernel which has its center at $z$ and the same size as $G$. In this way we obtain a scalar element by convolution. Note that the second derivatives of the Gaussian shape function are conventionally used for reducing the noise while reinforcing the response of the signal with a specific standard deviation (in this case $\sigma_f$) \cite{Voorn2013}. 
Thus, the elements in eq.~\ref{eq:Hessian_element} represent an enhanced intensity response of the aperture of the fracture in the original input image.

Defining the local structure at pixel $z$ with the help of second order partial derivatives  could be done by calculating the eigenvalues of the obtained Hessian matrix. In this case, since $\frac{\partial^2}{\partial{x}\partial{y}}I_f$ and $\frac{\partial^2}{\partial{y}\partial{x}}I_f$ are identical, the matrix is symmetric with real numbers thus, the eigenvalues of the matrix can be assured to be real numbers and its eigenvectors have to be orthogonal \cite{GoluVanl96}.

By defining the eigenvalues as $\lambda_1$ and $\lambda_2$, the line-like shape curvature have a relation of $|\lambda_1| \approx 0, |\lambda_2| \gg 0$ or vice versa. This is because each eigenvalue states an amount of gradient change at the selected pixel to each eigenvector direction (orthogonal when $\lambda_1 \neq \lambda_2$). Then, in the case of line-like shape (long and thin), one must have a very small value of $|\lambda_1|$ while the other has relatively bigger value of $|\lambda_2|$. The author proposed to replace the intensity value at pixel $z$ to $|\lambda_2|$ under the above mentioned line-like shape condition and to zero otherwise. Finally, the maximum responses from the output images were accumulated in a range of different $\sigma_f$ to maximize the effects for various fracture apertures.

Based on this scheme, the multi-scaled Sato filtering method was applied on the noise reduced data with the \texttt{sato} function from python \textit{skimage} library. Since the output of this method is an enhanced response of a string-like shape while minimizing the responses from the other shapes of structures, a further binarization method has to be applied to the output of the Sato method in order to obtain logical type of results. Here, we adopted the local threshold method (cf. \ref{sec:local_threshold}) which was able to effectively distinguish the enhanced responses out of the output. As we mentioned in \ref{sec:local_threshold}, before applying the binarization method, the outer part of the scanned data was eliminated since the contrast between the outer and inner part is bigger than the contrast of the fractures in the data.

After applying the binarization scheme, the remaining artifacts were eliminated with the use of morphological schemes such as \texttt{erosion} and \texttt{remove\_small\_object} which are also supported in the Python \textit{skimage} library.

\section{Active contouring}\label{appendix:Active_contouring}
\renewcommand{\theequation}{D.\arabic{equation}}
By employing the method which was introduced in the work of \cite{Chan2001}, we were able to obtain finer profiles from the segmented images. The method requires an initial mask which is a roughly segmented binary image and preferably contains most of the features of interest. Thus, we adopted the result of LT (cf. \ref{sec:local_threshold}) as the initial mask. From the given initial mask and the corresponding original image, we proceeded with the method to obtain a segmented image. This could be done by minimizing the differences between the mean intensities of in-/out areas, defined by the boundaries of the initial mask $C$, and the intensities from the original images $I(x,y)$ based on a modified version of the \textit{Mumford-Shah} functional \cite{Mumford1989},
\begin{linenomath*} 
\begin{equation}\label{eq:chan_vese_governing_equation}
     F(c_1,c_2,C) = \mu L(C) + \nu A(in(C)) + \lambda_1\int\limits_{in(C)}|I(x,y) - c_1|^2\,\text{d}x\text{d}y + \lambda_2\int\limits_{out(C)}|I(x,y) - c_2|^2\,\text{d}x\text{d}y,
\end{equation}
\end{linenomath*}
where $F$ is the functional of $c_1$ and $c_2$ which are the average intensities within the region of in-/outside of $C$, and $L$ is the length and $A$ is the inside area of $C$. The $\mu$, $\nu$, $\lambda_1$ and $\lambda_2$ are input parameters. 

In order to derive the numerical description within the common domain $(x,y) \in \Omega$, the level set scheme, which is able to describe the boundaries of a feature of interest and its area by introducing higher dimensional manifold \cite{Osher2003}, was combined in this relation. By following this scheme, the boundaries of the initial mask $C$ and inner-/outer areas of it can be defined such as,
\begin{linenomath*}
\begin{equation}
    \begin{cases}
    C : \Phi(x,y) =0, \\
    in(C) : \Phi(x,y) > 0, \\
    out(C) : \Phi(x,y) < 0,
    \end{cases}
\end{equation}
\end{linenomath*}
where $\Phi(x,y)$ is a Lipschitz continuous function to assure that it has a unique solution for a given case.

With the help of the scheme and a Heavyside step function $H$, eq. ~\ref{eq:chan_vese_governing_equation} could be rewritten within the common domain $\Omega$ as follows, 
\begin{linenomath*}
\begin{equation}
\begin{split}
    F(c_1,c_2,\Phi) =& \mu \int\limits_{\Omega}|\nabla{H(\Phi(x,y))}| \,\text{d}x\text{d}y +\nu \int\limits_{\Omega}H(\Phi(x,y)) \,\text{d}x\text{d}y\\
    &+\lambda_1\int\limits_{\Omega}|I(x,y) - c_1|^2H(\Phi(x,y))\,\text{d}x\text{d}y +\lambda_2\int\limits_{\Omega}|I(x,y) - c_2|^2 (1-H(\Phi(x,y))) \,\text{d}x\text{d}y.
\end{split}    
\end{equation}
\end{linenomath*}
For fixed $\Phi$, the intensity averages of interior $c_1$ and exterior $c_2$ of the contour could be obtained such as,
\begin{linenomath*}
\begin{equation}
    c_1(\Phi) = \frac{\int_{\Omega}{I(x,y)H(\Phi(x,y))}\,\text{d}x\text{d}y}{\int_{\Omega}H(\Phi(x,y))\,\text{d}x\text{d}y},
\end{equation}
\end{linenomath*}
and
\begin{linenomath*}
\begin{equation}
    c_2(\Phi) = \frac{\int_{\Omega}{I(x,y)(1-H(\Phi(x,y)))}\,\text{d}x\text{d}y}{\int_{\Omega}(1-H(\Phi(x,y)))\,\text{d}x\text{d}y}.
\end{equation}
\end{linenomath*}
Based on these relations, in order to minimize $F$, the authors used the following scheme by relating it with an Euler-Lagrange equation. By applying the artificial time $t \geq 0$, the descending direction of $\Phi$ is such as,
\begin{linenomath*} 
\begin{equation}
\begin{split}
    \frac{\partial{\Phi}}{\partial{t}} = \delta(\Phi)\left[\mu\, \text{div}\left(\frac{\text{grad}({\Phi})}{|\text{grad}({\Phi})|}\right) - \nu - \lambda_1(I(x,y) - c_1)^2 + \lambda_2(I(x,y)-c_2)^2 \right],\\
\end{split}
\end{equation}
\end{linenomath*}
where $\delta$ is the dirac delta function which is the first derivative of the Heavyside function $H$ in one-dimensional form thus, only the zero level-set part could be considered. From the initial contour $\Phi(x,y,t=0)$, a new contour was able to be defined by using the above relation till it reached to stationary state via iteration of the artificial time $t$.

Based on the theory, the method was applied on the original data which had not been dealt with the noise reduction technique. Additionally, the same masking criteria which were applied before the local threshold method ~\ref{sec:local_threshold} were employed before using the method. This is because the method is not able to cope with the image data which contain a large contrast between the inner and outer part of the sample, as mentioned before. Thus, by covering the exterior part with a manually extracted mean intensity of the non-fracture matrix, an effective segmentation of the fracture was able to be performed. Furthermore, in order to capture faint fractures which had a lower contrast than the other fractures, the method was applied on the cropped images with a size of (400 $\times$ 400 voxels). These segmented small tiles were merged later into the original size of the image.(See Figure~\ref{fig:xy_plain})

\section{Splitting, Training, Merging for the U-net model}\label{appendix:machinelearning}
\begin{figure}[!h]
    \centering
    \includegraphics[width=0.8\textwidth]{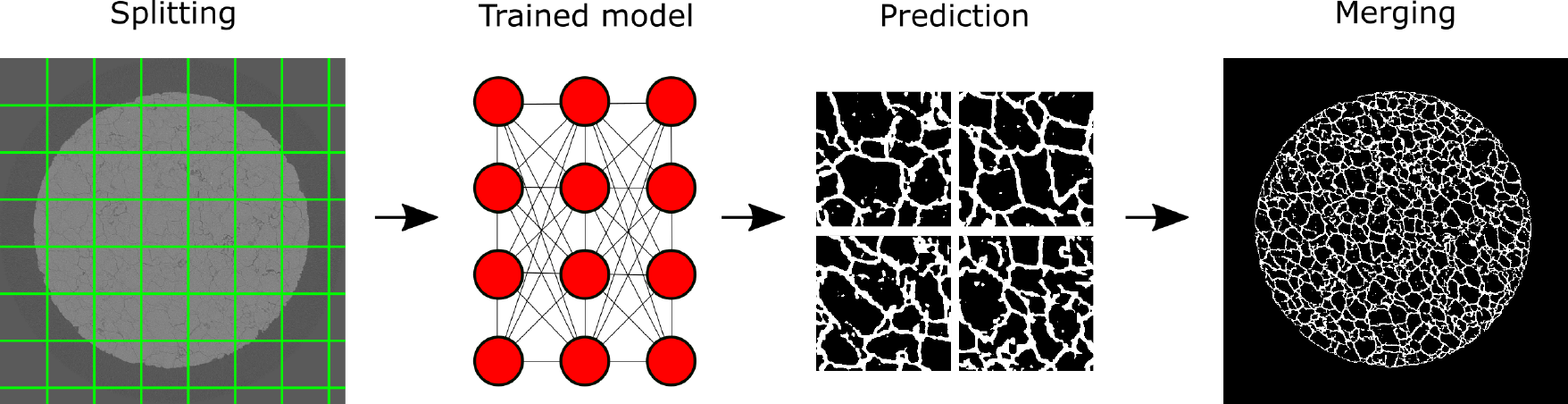}
    \caption{The workflow of the machine learning-based model. The green grid indicates the edges of the diced tiles (400 $\times$ 400 voxels) from the original image (2940 $\times$ 2940 voxels). After inputting each fragment into the trained model, we obtained the corresponding predictions. By applying the threshold ($\approx$ 0.5 in the most cases) obtained by the Otsu method the tiles were binarized and merged to shape the original size of the image.}
    \label{fig:workflow_ml_based}
\end{figure}

\renewcommand{\theequation}{E.\arabic{equation}}
The model makes use of repeating down-scaling of the input image with the help of max-pooling layers, and up-scaling with a de-convolutional layer. Additionally, before and after each of the up-/down-scaling layers, the convolutional layers which extract the feature maps were used with an activation function which introduced a non-linearity into the model. Each of the extracted and down-scaled features were concatenated to the same size of the up-scaled features, in order to force the output pixels to be located at reasonable locations (see Figure \ref{fig:2DU-net_model}). The workflow of data processing is described in Figure \ref{fig:workflow_ml_based}, and the same principle holds for the Random Forest scheme.

\subsection{Splitting} \label{sec:Splitting}
Due to the in-house computer's memory limitations, the big size of the complete data-set could not be dealt with the adopted machine learning model. This was because of the data which would be used for training and the model itself had to be allocated in the random-access memory RAM thus, a training data-set which exceeded its memory limitation could not be applied. With the given information, the size of the applicable data for training was decided by means of non-/trainable parameters, induced by the number and size of the layers within the model. The size of the images and the batch size of the training data indicated the number of training data per one iteration of training (See~\ref{sec:Training}). In our application moreover, since we wanted to take a benefit from the GPU, which could boost the training and processing time, the limitation of usable memory was even more restricted within the memory size of a GPU, which is normally smaller than the RAM and is not easily able to be expanded.

Consequently, due to these facts, the original size of 2D image data was split into small tiles. The original image size of 2940 $\times$ 2940 pixels was cropped into multiple 400 $\times$ 400 smaller images. We already knew that this could potentially induce some disconnection of the segmented fractures along the tiles, since the segmentation accuracy of the U-net model is low at the edges of given data due to the inevitable information loss during up-/down-scaling. In order to overcome and compensate this drawback, the overlapping scheme was applied, splitting the tiles to have 72 pixels of overlapping region at their borders. In this manner, 81 tiles were produced per 1 slice of a 2D original image. In addition, the format of the images which was ``unsigned int 16'' originally, was normalized into ``unsigned int 8'' after cutting the evaluated mean minimum and maximum intensities. This helped to reduce the memory demand caused by training data while preventing an over-simplification of the inherent histogram profiles.

\subsection{Training} \label{sec:Training}

Training was performed with the training data-set mapping the cropped original image and GT. Only few data (60 slices) of scanned data (where the data was located at the top part of the sample) was used as training data-set. In case of the training of the U-net model, this number of slices corresponded to 4860 cropped tiles. Among the data-set, 80 percent of the data (3888 tiles) were applied for updating the internal coefficients of the model, and the rest of the data (972 tiles) were used to evaluate the performance of the trained model by comparison between the predictions and GT.

For the sake of accuracy, the data augmentation technique was applied on the training data-set. This allowed us to enrich the training data-set by employing a modification to the data thus, the model could be trained with sufficient data of different variations. 
Consequently, the model would be trained with more trainable data. This contributed in the prevention of over-fitting, which made the model to be capable of dealing only with a specific case. In our application, we varied the brightness of the training data. Thus, the model was able to be trained by data with variation.  This was necessary to get good predictions from all cropped tiles.

The batch size in machine learning defines how many of training data to be used for updating the inner parameters of the model during training. This implies that the batch size would directly affect the memory usage since the defined numbers of training data would be held on the memory. Therefore, in this case, this number was selected carefully as 5 so as not to reach our GPU memory limitation. The steps per epoch was set at 777, estimated based on the following relation: $\approx \frac{\text{training data} : 3888}{\text{batch size} : 5}$ in a way that all the provided trainable data could be applied on each epoch. Finally, the model was trained with 100 epochs so that the segmentation accuracy could be enhanced along the iterations.

We adopted ``binary cross entropy'' loss function in order to deal with the binarization of image (classification of non-/fractures). The loss function $L$ is defined such as,
\begin{linenomath*}
\begin{equation}\label{eq:binary_cross_entropy}
    L = -g_t \log{(p)} - (1-g_t)\log{(1-p)},
\end{equation}
\end{linenomath*}
where $g_t$ is the given truth, and $p$ is the predicted probability of the model. Thus, the model was updated in a way that the difference between truths and predictions could be narrowed down. The ``Adam'' optimizer \cite{Kingma2015} was used for this optimization. The adopted fixed learning rate was 1e-4.

\subsection{Merging}\label{sec:Merging}
After training the model by following the procedure mentioned earlier \ref{sec:Training}, the model was able to predict the non-/fractures within the given tiles. In order to obtain the same size of output with the original image, we performed this merging procedure which was basically placing every prediction tile at their corresponding locations, after collecting the predictions from the entire data-set from the model. Since we applied the overlapping scheme \ref{sec:Splitting} in order to prevent disconnections between tiles, 36 pixels at the edges were dropped for each tile which were considered as inaccurate predictions before merging. After the merging, the predictions which were in the range of 0 to 1 due to the output layer of the training model were binarized with the Otsu method \cite{NobuyukiOtsu1979} for the final segmentation output. The pixels which were classified as zeros in the predictions were excluded from this binarization by assuming that these pixels were certainly non-fracture.


\noappendix       




\appendixfigures  

\appendixtables   


\authorcontribution{DL: implementation and performance of image analyse/segmentation, data analysis, original draft preparation. NK: original draft preparation, designed research. MR: data curation, XRCT scanning, original draft preparation. HS: original draft preparation, designed research.} 

\competinginterests{The authors declare that they have no conflict of interest} 


\begin{acknowledgements}
This work was funded by the Deutsche Forschungsgemeinschaft (DFG, German Research Foundation) – Project Number 327154368 – SFB 1313.  Holger Steeb is thanking the Deutsche
Forschungsgemeinschaft (DFG) for supporting this work by funding EXC 2075-390740016 under Germany’s Excellence Strategy.
\end{acknowledgements}



\bibliographystyle{copernicus}
\bibliography{reference.bib}

\end{document}